\documentclass[twocolumn, draft]{svjour3}\usepackage[]{graphicx}\usepackage[]{color}
\makeatletter
\def\maxwidth{ %
  \ifdim\Gin@nat@width>\linewidth
    \linewidth
  \else
    \Gin@nat@width
  \fi
}
\makeatother

\definecolor{fgcolor}{rgb}{0.345, 0.345, 0.345}

\usepackage{framed}
\makeatletter
 {\par\unskip\endMakeFramed%
 \at@end@of@kframe}
\makeatother

\definecolor{shadecolor}{rgb}{.97, .97, .97}
\definecolor{messagecolor}{rgb}{0, 0, 0}
\definecolor{warningcolor}{rgb}{1, 0, 1}
\definecolor{errorcolor}{rgb}{1, 0, 0}
\newenvironment{knitrout}{}{} 

\usepackage{alltt}          
\smartqed  
\usepackage{amsmath}
\usepackage{graphicx,psfrag,epsf}
\usepackage{enumerate}
\usepackage{natbib}
\usepackage{url} 

\usepackage{alltt}
\usepackage{amsmath}

\usepackage{enumerate}
\usepackage{bbm}
\usepackage{bm}
\usepackage{amssymb}
\usepackage{amsmath}
\usepackage{calligra}
\IfFileExists{upquote.sty}{\usepackage{upquote}}{}
\begin{document}

\title{Inference for $L_2$-Boosting
}


\author{David R\"ugamer         \and
        Sonja Greven 
}


\institute{David R\"ugamer \at
              Department of Statistics, LMU Munich\\
              Ludwigstr. 33, 80539, Munich, Germany \\
              \email{david.ruegamer@stat.uni-muenchen.de}           
           \and
           Sonja Greven \at
              Chair of Statistics, School of Business and Economics,\\ Humboldt University of Berlin\\
              Unter den Linden 6, 10117, Berlin, Germany \\
              \email{sonja.greven@hu-berlin.de}
}

\date{Received: date / Accepted: date}

\maketitle

\def\spacingset#1{\renewcommand{\baselinestretch}%
{#1}\small\normalsize} \spacingset{1}


%

\maketitle

\bigskip
\begin{abstract}
We propose a statistical inference framework for the component-wise functional gradient descent algorithm (CFGD) under normality assumption for model errors, also known as $L_2$-Boosting. The CFGD is one of the most versatile tools to analyze data, because it scales well to high-dimensional data sets, allows for a very flexible definition of additive regression models and incorporates inbuilt variable selection. 
Due to the variable selection, we build on recent proposals for post-selection inference. However, the iterative nature of component-wise boosting, which can repeatedly select the same component to update, 
necessitates adaptations and extensions to existing approaches. We propose tests and confidence intervals for linear, grouped and penalized additive model components selected by $L_2$-Boosting. Our concepts also transfer to slow-learning algorithms more generally, and to other selection techniques which restrict the response space to more complex sets than polyhedra.
We apply our framework to an additive model for sales prices of residential apartments and investigate the properties of our concepts in simulation studies. 
\keywords{Bootstrap \and Functional Gradient Descent Boosting \and Post-Selection Inference \and Selective Inference \and Slow Learner}
\end{abstract}

\newpage
\spacingset{1.45} 

\section{Introduction} \label{sec:intro}

We propose statistical inference techniques for the com\-po\-nent-wise functional gradient descent algorithm \citep[CFGD; see, e.g.,][]{Hothorn.2010}. CFGD emerged from the field of machine learning \citep[c.f.][]{Friedman.2001}, but has since also become an algorithm used to estimate statistical models \citep[see, e.g.,][]{
Mayr.2017, Melcher.2017, 
Ruegamer.2018a, Brockhaus.2018}. The CFGD is an iterative procedure, which performs model updates in the direction of the steepest decent with respect to a chosen loss function and, in contrast to other gradient boosting algorithms, only adds one single additive term (base-learner) to the model in each iteration. The algorithm is typically used in applications, where the goal is to obtain variable selection, similar to the Lasso but with the additional flexibility to estimate any type of additive regression model. The variable selection is implicitly given by the component-wise updates in combination with early stopping of the algorithm to avoid overfitting. Examples for additive regression models, which are based on the CFGD fitting procedure, are generalized additive models or functional regression models, potentially in combination with a non-normal response. In some applications such as complex function-on-function regression \citep[see, e.g.,][]{Ruegamer.2018a}, the CFGD also facilitates the estimation and modular extension of a model, which cannot be fitted with other standard software packages. The main difference and advantage lies in its component-wise fitting nature, iteratively fitting only one additive term to the response at a time and thereby allowing for a large number of covariates with manageable computational costs. 
A commonly used and well studied special CFGD algorithm is \emph{$L_2$-Boosting} \citep{Buehlmann.2003}. 
No general inferential concepts in the sense of classical statistical inference have been proposed for $L_2$-Boosting yet. Ad-hoc solutions such as a non-parametric bootstrap are often used to quantify the variability of boosting estimates \citep[see e.g.][]{Brockhaus.2015, Ruegamer.2018a}, although this does not lead to confidence intervals with proper coverage. 
In many research areas uncertainty quantification is indispensable. 
We propose a framework to conduct valid inference for regression coefficients in models fitted with $L_2$-Boosting by conditioning on the selected covariates. 
We build on recent research findings on \textit{selective inference}, which transfer classical statistical inference to algorithms with preceding selection of model terms, as is also the case for CFGD algorithms. 

%

Standard inference is invalid after model selection, as mentioned by many authors throughout the last few de\-cades 
\citep[see, e.g.,][]{Berk.2013}, and a suitable inference framework is required. Different approaches for inference in high-dimensional regression models have emerged over the past few years, including data splitting \citep{Wasserman.2009} and 
more recently, post-selection inference \citep[PoSI;][]{Berk.2013} for valid statistical inference after arbitrary selection procedures. In this paper, classical statistical inference refers to inference concepts usually applied to assess uncertainty in regression models that do not account for a preceding selection or model choice in any sense, but treat the empirically selected model as given a priori. Invalidity of classical statistical inference methods after model selection can, in part, be explained by the fact, that the data generating process of the response will usually not yield only one specific but different selected models for a given model selection procedure for different realizations $\bm{y}$ of the response $\bm{Y} \in \mathcal{Y}$. From a geometrical point of view, different subspaces of the space $\mathcal{Y}$ will thus yield different selected models. When conditioning on a specific model for inference statements, this can be regarded as conditioning on a subspace of $\mathcal{Y}$ for inference. Classical inference methods, however, assume that the model is known prior to the analysis and hence that $\mathcal{Y}$ is not restricted. A restriction of the space of $\bm{Y}$ in turn results in a restriction of the distribution for $\hat{\bm{\beta}}$, which, if not accounted for, yields to over optimistic inference statements for the estimated parameters. We focus on \textit{selective inference}, which provides inference statements conditional on the observed model selection. Similar to data splitting, selective inference separates the information in the data used for model selection from the information used to infer about parameters post model selection. In contrast to the original PoSI idea of providing simultaneous inference for every possible model selection, selective inference is designed to yield less conservative inference statements.

\citet{Fithian.2014} have developed a general theory for selective inference in exponential family models following any type of selection mechanism. Additionally, different explicit selective inference frameworks have been derived for several selection methods (\citealp[see e.g.][for selective inference after Lasso selection]{Lee.2016} or \citealp[][for selective inference after likelihood- and test-based model selection]{Ruegamer.2018b}). Recent work, which we adapt and extend, aims for valid inference in forward stepwise regression \citep{Tibshirani.2016, Loftus.2014, Loftus.2015a}. 


Compared to these approaches, inference for $L_2$-Boosting carries additional challenges due to an iterative procedure that can repeatedly select the same model term. We also extend our approach to allow for non-linear covariate effects, in contrast to existing approaches.

Our contributions are as follows: 
1. We explicitly derive the space restriction of the response given by the $L_2$-Boosting path and thereby allow for inference as proposed in \citet{Tibshirani.2016}.
2. We propose a new and more powerful conditional inference concept for $L_2$-Boosting by conditioning only on the \underline{set of selected variables}, i.e., on a set of possible selection paths. This idea can also be used for other slow learning algorithms that would require conditioning on additional quantities, with a resulting potential loss in power, to obtain an analytic representation of the inference space. For additive model structures we consider slow learners as algorithms that can repeatedly use the same additive term to gradually update a model, often by adding or deleting one covariate respectively from the model at a time. The CFGD or Forward Stagewise Regression are known examples exhibiting this behaviour. Another example is the Lasso, where an analytic representation of the inference space only becomes feasible after additionally conditioning on a list of signs and the order of variables selected.
3. We compute p-values and (two-sided) confidence intervals by Monte Carlo approximation following the results of \citet{Tibshirani.2016} and \citet{Yang.2016}. This circumvents an explicit mathematical representation of the space the test statistic is truncated to.  We refine their approach with a sampling routine that is more efficient in our setting. This approach is more generally applicable whenever the model of interest is of additive nature and the response variable is assumed to be normally distributed.
4. We extend the inference concept to account for cross-validation, stability selection \citep{Shah.2013} and similar sub-sampling methods. 
5. We further extend the approach to models including $L_2$-penalized additive effects, such as smooth effects. 


Below, we summarize the $L_2$-Boosting algorithm in Section \ref{sec:boosting} and the concept of selective inference for sequential regression procedures 
in Section \ref{sec:si_selproc}. We discuss the challenges accompanying an inference framework for $L_2$-Boosting and our proposed solutions in Section \ref{sec:adaption}. Section \ref{sec:simulation} presents simulation results. Section~\ref{sec:application}  analyzes sales prices of real estate apartments in Tehran using our new approach. We discuss limitations and further extensions of the approach in Section \ref{sec:extensions}. An add-on \textsf{R}-package to the model-based boosting \textsf{R} package \texttt{mboost} is available at \url{https://github.com/davidruegamer/iboost} and can be used to conduct inference for boosted models and to reproduce the results of sections \ref{sec:simulation} and \ref{sec:application}. Further simulation and application results as well as a code to reproduce the simulation results are given in the Supplementary Material.

\section{$L_2$-Boosting} \label{sec:boosting}

Let $\bm{X} \in \mathbb{R}^{n \times p}$ be a fixed set of covariates and $\bm{y}$ a realization of the random response variable $\bm{Y} \in \mathbb{R}^n$. The goal of component-wise gradient boosting \citep[see, e.g.,][]{Buehlmann.2007} is to minimize a loss function $\ell(\cdot,\bm{y})$ for the given realization $\bm{y}$ with respect to an additive model $\bm{f} := \sum_{j=1}^J g_j(\bm{X}_j)$, where function evaluations of $g_j$ are evaluated row-wise. The functions $g_j(\cdot)$, the so called base-learners, are defined for column subsets $\bm{X}_j \in \mathbb{R}^{n \times p_j}$ of $\bm{X}$ with $1\leq p_j \leq p$ and can be fitted to some vector $\bm{u}^{(m)} \in \mathbb{R}^n$, which yields $\hat{\bm{g}}_j^{(m)}$ as estimate for $g_j(\bm{X}_j)$. We estimate $\bm{f}$ by $\hat{\bm{f}}$ using the component-wise functional gradient descent algorithm:

\begin{enumerate}[(1)]
\setlength{\itemsep}{0pt}
  \setlength{\parskip}{0pt}
\item Initialize an offset value $\hat{\bm{f}}^{(0)} \in \mathbb{R}^n$. If $\bm{y}$ is centered, a natural choice is $\hat{\bm{f}}^{(0)} = (0, \ldots, 0)^\top$. Define $m = 0$.
\item Do the following for $m = 1, \ldots, m_{stop}$:
\begin{itemize}
    \item[(2.1)] Compute the pseudo-residuals $\bm{u}^{(m)} \in \mathbb{R}^n$ of step $m$ as $\bm{u}^{(m)} = - \left. \frac{\partial}{\partial \bm{f}} \ell (\bm{f},\bm{y})\right|_{\bm{f}=\hat{\bm{f}}^{(m-1)}}$. 
    \item[(2.2)] Approximate the negative gradient vector $\bm{u}^{(m)}$ with $\hat{\bm{g}}_j^{(m)}$ by fitting each of the base-learners ${g}_j(\cdot), j=1,\ldots,J$ to the pseudo-residuals and find the base-learner $j^{(m)}$, for which $j^{(m)} = \text{argmin}_{1 \leq j \leq J}\, || \bm{u}^{(m)} - \hat{\bm{g}}_j^{(m)} ||_2^2\,$ holds. 
    \item[(2.3)] Update $\hat{\bm{f}}^{(m)} = \hat{\bm{f}}^{(m-1)} + \nu \cdot \hat{\bm{g}}_{j^{(m)}}^{(m)}$, where $\nu \in (0,1]$ is the so called \emph{step-length} or \emph{learning rate} and usually fixed to some sufficiently small value such as $0.1$ or $0.01$ \citep{Buehlmann.2007}.
    \end{itemize}
    \end{enumerate}

\noindent When defining $\ell(\bm{f},\bm{y}) = \frac{1}{2} || \bm{y} - \bm{f}||_2^2$ with quadratic $L_2$-Norm $|| \cdot ||_2^2$, 
$L_2$-Boosting is obtained, which corresponds to mean regression using the model $\mathbb{E}(\bm{Y}|\bm{X}) = \sum_{j=1}^J g_j(\bm{X}_j)$. The vector $\bm{u}^{(m)}$ then corresponds to the residuals $\bm{y} - \hat{\bm{f}}^{(m-1)}$. 
In the framework of additive regression models, each base-learner $g_j(\cdot)$ constitutes a partial effect and is represented as a linear effect of a covariate or of a basis evaluated at that covariate vector, i.e., ${\bm{g}}_j(\bm{X}_j) = \bm{X}_j {\bm{\beta}}_j$.  
The coefficient $\bm{\beta}_j$ is estimated using ordinary or penalized least squares. 
The model fit $\hat{\bm{g}}_j^{(m)}$ of each base-learner in the $m$th step is therefore given by $\hat{\bm{g}}^{(m)}_{j} = \bm{H}_j \bm{u}^{(m)} = \bm{X}_j (\bm{X}_j^\top \bm{X}_j + \lambda_j \bm{D}_j)^{-1} \bm{X}_j^\top \bm{u}^{(m)}$, where the hat matrix $\bm{H}_j$ is defined by the corresponding design matrix $\bm{X}_j$, a penalty matrix $\bm{D}_j$ and a pre-specified smoothing parameter $\lambda_j \geq 0$ controlling the penalization. 
As only one base-learner is chosen in each iteration, the final effective degrees of freedom of the $j$th base-learner depend on the number of selections.

$L_2$-Boosting scales well to large data sets due to its component-wise fitting nature and is particularly suited for the estimation of structured additive regression models. 
It has the additional advantage of being able to handle $n<p$-settings and conducting variable selection, as not all $J$ model terms are necessarily selected in at least one iteration. However, variable selection has to be accounted for when constructing uncertainty measures for regression coefficients, as it restricts the space of $\bm{Y}$ and thus of the estimated parameters. 

\section{Selective Inference} \label{sec:si_selproc}


\subsection{Considered Setup}

Let $\bm{Y} = \bm{\mu} + \bm{\varepsilon}$ with $\bm{\varepsilon} \sim \mathcal{N}(\bm{0}, \sigma^2 \bm{I}_n)$ and $n$-dimensional identity matrix $\bm{I}_n$. Furthermore, assume that $\sigma^2$ is known and $\bm{\mu}$ is an unknown parameter of interest. We do not assume any true linear relationship between $\bm{\mu}$ and covariates, but estimate $\bm{\mu}$ with an additive ``working model'' based on fixed covariates $\bm{X} \in \mathbb{R}^{n \times p}$ with $p$ potentially exceeding $n$. Furthermore, define the selection procedure or selection event $$\mathcal{S}: \mathbb{R}^n \to \mathcal{P}(\lbrace 1,\ldots,p \rbrace), \bm{y} \mapsto \mathcal{S}(\bm{y})$$ with power set function $\mathcal{P}(\cdot)$. For the given realization $\bm{y}$ of $\bm{Y}$, we denote $\mathcal{S}(\bm{y}) =: \mathcal{A}$, for which we assume $|\mathcal{A}| \leq n$. 

We focus on estimating the best linear projection of $\bm{\mu}$ into the space spanned by the variables given by $\mathcal{A}$ after model selection and making uncertainty statements about any direction of this projection, i.e., the significance of any linear covariate effect, given the selected model $\mathcal{A}$. We therefore run the selection procedure defined by $\mathcal{S}$, select the subset $\bm{X}_\mathcal{A}$ of $\bm{X}$ defined by the selected column indices $\mathcal{S}(\bm{y}) = \mathcal{A}$ and estimate regression coefficients $\bm{\beta}_\mathcal{A}$ by projecting $\bm{y}$ into the linear subspace ${\bm{W}}_\mathcal{A} \subseteq \mathbb{R}^n$ spanned by the columns of $\bm{X}_\mathcal{A}$. Our inference goal is to test one entry $\beta_j$ in $\bm{\beta}_\mathcal{A}$, i.e., $$H_0: \beta_j = \beta_{j,0},$$ conditional on the selected model, which is equivalent to testing 
\begin{equation}
H_0: \bm{v}^\top \bm{\mu} := \bm{e}_j^T (\bm{X}_\mathcal{A}^\top \bm{X}_\mathcal{A})^{-1} \bm{X}_\mathcal{A}^\top \bm{\mu} = \beta_{j,0}  \label{eq:testvecPH}
\end{equation}
with $\bm{e}_j$ the unit vector selecting  $j\in \mathcal{A}$ \citep[see, e.g.,][]{Tibshirani.2016}. Without selection, (\ref{eq:testvecPH}) can be tested using $\tilde{R} := \bm{v}^\top\bm{Y}$, which follows a normal distribution with expectation $\tilde{\rho} = \bm{v}^\top \bm{\mu}$ and variance $\sigma^2 \bm{v}^\top \bm{v}$ under the null. However, after model selection, the space of $\bm{Y}$ is restricted to $\mathcal{G} = \{\bm{y}: \mathcal{S}(\bm{y}) = \mathcal{A} \}$, which we call the \emph{inference region}. Many of the proposed methods for selective inference then describe this space restriction mathematically and derive the distribution of $\bm{v}^\top\bm{Y} \,|\, \bm{Y} \in \mathcal{G}$. 

Let $\bm{P}_{\bm{W}}$ generally be the projection onto a linear subspace $\text{span}(\bm{W}) \subset \mathbb{R}^n$ defined by some $\bm{W} \in \mathbb{R}^{n \times w},\allowbreak w \in \mathbb{N}$, and $\bm{P}^\bot_{\bm{W}}$ be the projection onto the orthogonal complement of this linear subspace. Furthermore, define the direction of $\bm{P}_{\bm{W}}\bm{y}$ as the unit vector $\text{dir}_{\bm{W}}(\bm{y}) = \frac{\bm{P}_{\bm{W}} \bm{y}}{||\bm{P}_{\bm{W}} \bm{y}||_2}$. 

We now shortly review three approaches to selective inference derived for a similar setup and build on these ideas in Section~\ref{sec:adaption}.

\subsection{Existing Approaches for Other Procedures} \label{sec:infph}

For sequential regression procedures such as Forward Stepwise Regression ($FSR$) or the Least Angle Regression 
\citep[$LAR$, ][]{Efron.2004}, \citet{Tibshirani.2016} characterize the restricted region of the on-going selection mechanism as a polyhedral set 
$\mathcal{G}= \{ \bm{y}: \bm{\Gamma} \bm{y} \geq \bm{b} \}$ with $\bm{\Gamma} \in \mathbb{R}^{\iota \times n}$, $\bm{b} \in \mathbb{R}^\iota$ for some $\iota \in \mathbb{N}$ and an inequality $\geq$ which is to be interpreted componentwise. 

By additionally conditioning on the realization $\bm{z}$ of $\bm{Z} = \bm{P}_{\bm{v}}^\bot \bm{Y}$ as well as on a list of signs for each step similar to those defined in (\ref{eq:boostingPH}) and which will be explained in Section~\ref{sec:adaption},
$\tilde{R}$ follows a truncated Gaussian distribution $\mathcal{N}(\tilde{\rho}, \sigma^2 \bm{v}^\top \bm{v})$ with analytically describable truncation limits $\mathcal{V}^{lo} = \mathcal{V}^{lo}(\bm{z}), \mathcal{V}^{up} = \mathcal{V}^{up}(\bm{z})$ \citep[see][]{Lee.2016}. Let $F^{[\mathcal{V}^{lo},\mathcal{V}^{up}]}_{\tilde{\rho}, \sigma^2 \bm{v}^\top \bm{v}}(\tilde{R})$ denote the cumulative distribution function of this truncated normal distribution evaluated at $\tilde{R}$. Then, for $H_0: \tilde{\rho} \leq 0$ vs. $H_1: \tilde{\rho} > 0$, the test statistic $T = 1 - F^{[\mathcal{V}^{lo},\mathcal{V}^{up}]}_{0, \sigma^2 \bm{v}^\top \bm{v}}(\tilde{R})$ is a valid conditional p-value, conditional on the polyhedral selection, as $\mathbb{P}_{H_0} (T \leq \alpha \mid \bm{\Gamma} \bm{Y} \geq \bm{b}) = \alpha$ for any $0 \leq \alpha \leq 1$. Two-sided p-values can be constructed as $T = 2 \cdot \min \left(F^{[\mathcal{V}^{lo},\mathcal{V}^{up}]}_{0, \sigma^2 \bm{v}^\top \bm{v}}(\tilde{R}), 1 - F^{[\mathcal{V}^{lo},\mathcal{V}^{up}]}_{0, \sigma^2 \bm{v}^\top \bm{v}}(\tilde{R}) \right)$ \citep{Tibshirani.2016}.

The characterization of the inference region as a polyhedral set, however, is only possible if the algorithmic decision in each selection step is a linear restriction on the space of $\bm{Y}$. 
\citet{Loftus.2015a} introduce a framework for inference after model selection procedures which can be described by affine inequalities, focusing on groups of variables. 
For testing the $j$th group variable coefficient $\bm{\beta}_{\mathcal{A},j} \in \mathbb{R}^{p_j}$ in the best linear approximation $\bm{\beta}_{\mathcal{A}} = \text{arg min } \allowbreak \mathbb{E} [\,|| \bm{Y} - \bm{X}_\mathcal{A} \bm{\beta} ||^2_2 \, ]$, 
\citet{Loftus.2015a, Yang.2016} rewrite the null hypothesis 
$\bm{\beta}_{\mathcal{A},j} = \bm{0}$ 
as $\bm{P}_{\bm{W}} \bm{\mu} = \bm{0}$ or  
\begin{equation}
H_0: \rho := ||\bm{P}_{\bm{W}} \bm{\mu}||_2 = 0 \label{eq:yanghyp}
\end{equation}
with ${\bm{W}} = \bm{P}^\bot_{\bm{X}_{\mathcal{A}\backslash j}} \bm{X}_j$, where $\bm{X}_{\mathcal{A}\backslash j}$ denotes $\bm{X}_\mathcal{A}$ without the $p_j$ columns corresponding to the $j$th group variable. 
 Under the null and when additionally conditioning on the direction $\text{dir}_{\bm{W}}(\bm{y})$, ${R} := ||\bm{P}_{\bm{W}} \bm{Y} ||_2$ follows a truncated $\chi$-distribution with analytically derivable limits. 
\citet{Yang.2016} note that $R$ and $\text{dir}_{\bm{W}}(\bm{y})$ are not independent for $\rho \neq 0$ and as a consequence, the $\chi_{}$-conditional distribution of $R$ as derived in \citet{Loftus.2015a} for (\ref{eq:yanghyp}) when $\rho = 0$ no longer holds for more general hypotheses, as relevant for the derivation of confidence intervals. 

\citet{Yang.2016} decompose $\bm{Y}$ as $R \cdot \text{dir}_{\bm{W}}(\bm{Y}) + \bm{P}^\bot_{\bm{W}} \bm{Y}$ and condition on $\text{dir}_{\bm{W}}(\bm{Y}) = \text{dir}_{\bm{W}}(\bm{y})$ as well as on $\bm{P}^\bot_{\bm{W}} \bm{Y} = \bm{P}^\bot_{\bm{W}} \bm{y}$. Then, the only variation left is in $R$ and the selection $\mathcal{A}$ can be equally written as $R \in \mathcal{R}_y$ with
\begin{equation}
\mathcal{R}_y = \left\{R > 0: \mathcal{S}(R \cdot \text{dir}_{\bm{W}}(\bm{y}) + \bm{P}^\bot_{\bm{W}} \bm{y}) = \mathcal{A} \right\}. \label{selinfEvent}
\end{equation}
The distribution of $R$ conditional on the selection, on $\text{dir}_{\bm{W}}(\bm{y})$ as well as on $\bm{P}^\bot_{\bm{W}} \bm{y}$, has a density proportional to
\begin{equation}
R^{w-1} \exp \left\{ - \frac{1}{2\sigma^2} (R^2 - 2R \cdot \langle \text{dir}_{\bm{W}}(\bm{y}), \bm{\mu} \rangle) \right\} \cdot \mathbbm{1} \{ R \in \mathcal{R}_y \} \label{eq:yangDens}
\end{equation}
with indicator function $\mathbbm{1}\{\cdot\}$. (\ref{eq:yangDens}) can be used to conduct inference on the inner product $\langle \text{dir}_{\bm{W}}(\bm{y}), \bm{\mu} \rangle$. As $\rho = ||\bm{P}_{\bm{W}} \bm{\mu}||_2 \geq \langle \text{dir}_{\bm{W}}(\bm{y}), \bm{\mu} \rangle$ holds, (\ref{eq:yangDens}) can also be used to construct a lower bound for the quantity of interest $\rho$.  




An explicit definition of the inference region is, however, not necessary. Theorem 1 in \citet{Yang.2016} states that, conditional on $\text{dir}_{\bm{W}}(\bm{y})$, $\bm{P}^\bot_{\bm{W}} \bm{y}$ and the selection event, inference can be conducted using the Uni\-form$[0,1]$ p-value $\varsigma(t_y)$ for $H_0: \langle \text{dir}_{\bm{W}}(\bm{y}), \bm{\mu} \rangle = t_y$ with 
\begin{equation}
\varsigma(t) = \frac{\int_{R \in \mathcal{R}_y, R > || \bm{P}_{\bm{W}} \bm{y}||_2} R^{w - 1} e^{-(R^2 - 2Rt)/2\sigma^2} \, \mathrm dR}{\int_{R \in \mathcal{R}_y} R^{w - 1} e^{-(R^2 - 2Rt)/2\sigma^2} \, \mathrm dR}. \label{eq:fFromYang}
\end{equation}
The authors note that (\ref{eq:fFromYang}) is equal to
\begin{equation}
\frac{\mathbb{E}_{R\sim\sigma \chi_{w}}(e^{Rt/\sigma^2} \cdot \mathbbm{1} \{R\in \mathcal{R}_y,  R > || \bm{P}_{\bm{W}} \bm{y}||_2 \})}{\mathbb{E}_{R\sim\sigma \chi_{w}}(e^{Rt/\sigma^2} \cdot \mathbbm{1} \{R\in \mathcal{R}_y \})}, \label{eq:eFromYang}
\end{equation}
which can be approximated by the ratio of empirical expectations computed with a large number of samples $r^b \sim \sigma \cdot \chi_{w}, b=1,\ldots,B$. To evaluate the argument of both expectations in (\ref{eq:eFromYang}) for some $r^b$, $r^b \in \mathcal{R}_y$ must be checked. Note that the only variation of $(\bm{Y} \mid \text{dir}_{\bm{W}}(\bm{y}), \bm{P}^\bot_{\bm{W}} \bm{y})$ is in $R$. Therefore, define $\bm{y}^b =  \bm{P}^\bot_{\bm{W}} \bm{y} + r^b \cdot \text{dir}_{\bm{W}}(\bm{y})$ and rerun the algorithm to check whether $\mathcal{S}(\bm{y}^b) = \mathcal{A}$, or equivalently, whether $r^b \in \mathcal{R}_y$. Drawing samples from the $\sigma \chi_w$-distribution is inefficient, however,  when $|| \bm{P}_{\bm{W}} \bm{y}||_2$ is far away from the null as then an excessively large number of samples is needed to obtain a good approximation of $\varsigma(t)$. \citet{Yang.2016} therefore suggest an importance sampling algorithm, which draws samples $r^b$ from a proposal distribution $\mathcal{F}_{prop}$ such as $\mathcal{N}(|| \bm{P}_{\bm{W}} \bm{y}||_2, \sigma^2)$ with density $f_{prop}$ and then approximates (\ref{eq:eFromYang}) by
\begin{equation}
\varsigma(t) \approx \hat{\varsigma}(t) = \frac{\sum_b w_b \cdot e^{r^b t/\sigma^2} \cdot \mathbbm{1} \{r^b \in \mathcal{R}_Y,  r^b > || \bm{P}_{\bm{W}} \bm{y}||_2 \}}{\sum_b w_b \cdot e^{r^b t/\sigma^2} \cdot \mathbbm{1} \{r^b \in \mathcal{R}_Y \}} \label{yangapprox}
\end{equation}
with sampling weights $w_b = f_{\sigma \chi_{w}}(r^b) / f_{prop}(r^b)$.

\section{Selective Inference concepts for $L_2$-Boosting} \label{sec:adaption}

We now propose selective inference concepts for $L_2$-Boosting. In Section~\ref{subsec:l2ph} we first derive a polyhedron representation of selection conditions in $L_2$-Boosting. After discussing the resulting inference framework based on existing concepts and its lack of power in Section~\ref{subsec:adaption1}, we propose an alternative concept for $L_2$-Boosting and similar slow learners, which can repeatedly select the same base-learners. Based on this idea, we derive a powerful inference framework for $L_2$-Boos\-ting with linear base-learners in Section~\ref{subsec:adaption2} and describe important extensions in Section~\ref{extensions}.

%


\subsection{Polyhedron representation-based inference for $L_2$-Boosting} \label{subsec:l2ph}

Consider $L_2$-Boosting using only linear base-learners, i.e., $\bm{D}_j = \bm{0}, \bm{X}_j \in \mathbb{R}^{n\times 1}\, \forall\,j$. Similar to \citet{Tibshirani.2016}, we can derive a polyhedron representation $\mathcal{G}= \{ \bm{y}: \bm{\Gamma} \bm{y} \geq \bm{b} \}$ \textbf{for the given selection path} $j^{(1)},\ldots,\allowbreak j^{(m_\text{stop})}$ of $L_2$-Boosting. 

The selection condition for the $m$th chosen base-learner 
\begin{equation}
\begin{split}
\quad &||(\bm{I}-\bm{H}_{j^{(m)}})\bm{u}^{(m)}||^2 \leq ||(\bm{I}-\bm{H}_{j})\bm{u}^{(m)}||^2\\
\Leftrightarrow \quad &\left( s_m \bm{X}_{j^{(m)}}^\top / ||\bm{X}_{j^{(m)}}||_2 \pm \bm{X}_j^\top/||\bm{X}_j||_2 \right) \bm{u}^{(m)} \geq 0, \label{eq:selcond}
\end{split}
\end{equation}
which holds $\forall j \neq j^{(m)}$ with $s_m = \text{sign}(\bm{X}_{j^{(m)}}^\top \bm{u}^{(m)})$, can be written as affine restriction on $\bm{y}$ by plugging the residual vector $\bm{u}^{(m)}$ of step $m$ as a function of $\bm{y}$ $$\bm{u}^{(m)} = \left[ \prod_{l=1}^{m-1} \left( \bm{I} - \nu \bm{H}_{j^{(m-l)}} \right) \right] \bm{y} =: \Upsilon^{(m)} \bm{y}$$ into (\ref{eq:selcond}). 
For a given selection path and list of signs $s_m, m=1,\ldots,m_\text{stop}$ this yields the polyhedron representation $\mathcal{G}$ with fixed $(2 \cdot (p-1) \cdot m_\text{stop}) \times n$ matrix $\bm{\Gamma}$ as stacked matrix of $n$-dimensional row vectors, where the rows $(\tilde{m} + 2(j - \omega(j))-1)$ and $(\tilde{m} + 2(j -\omega(j)))$ of $\bm{\Gamma}$ with $\tilde{m} = 2 \cdot (p-1) \cdot (m-1)$ and $\omega(j) = \mathbbm{1}\{j>j^{(m)}\}$ are given by 
\begin{equation}
\bigl( s_m \bm{X}_{j^{(m)}}^\top / ||\bm{X}_{j^{(m)}}||_2 \pm \bm{X}_{j}^\top / ||\bm{X}_{j}||_2 \bigr) \Upsilon^{(m)} \quad \forall\,j \neq j^{(m)}. \label{eq:boostingPH}
\end{equation}
As for other procedures described in the post-selection inference literature, this representation only holds if the columns of $\bm{X}$ are in general position, which however, is not a very stringent assumption \citep[see, e.g.,][Section 4]{Tibshirani.2016}. 

As the $L_2$-Boosting path results in a polyhedral set as space restriction for $\bm{Y}$, conditional on the list of signs, quantities of interest $\bm{v}^\top \bm{\mu}$ can be tested based on the conditional distribution of $\bm{v}^\top \bm{Y} \, | \, \bm{Y} \in \mathcal{G}$ as proposed by \citet{Tibshirani.2016}. To this end, we have to condition on the selection path. If we do not additionally condition on the list of signs, $\mathcal{G}$ is a union of polyhedra \citep[cf.][]{Lee.2016}. 

If group base-learners or base-learners with penalties are used, space restrictions no longer yield a polyhedron. Instead, affine inequalities can be used to obtain truncation limits analogous to \citet{Loftus.2015a, Ruegamer.2018b}.




\subsection{Choice of the Conditioning Event for Slow Learners} \label{subsec:adaption1}

For the selection approaches discussed in Section~\ref{sec:si_selproc}, conditioning on the selection path helps to derive the corresponding conditional distribution and, compared to conditioning on the selected model only, additionally conditions on the selection order of variables and their effect sign. 
For boosting and other slow learners that can repeatedly select the same base-learner, conditioning on the selection path and thus on variable selection decisions in each algorithmic step will result in an even larger loss of power. In fact, such a conditional inference will have almost no power in most practically relevant situations, as we show empirically for the polyhedron approach in the simulation section. In order to avoid excessive conditioning, we propose conditioning only on the \underline{set of selected covariates} (not on the selection order or the effect signs), i.e., on the selected statistical model. 

Conditioning only on the selected covariates, however, means that the mathematical description of the inference region becomes far more difficult. For $L_2$-Boos\-ting with linear base-learners, this would result in a union of not necessarily overlapping polyhedra for the different selection paths leading to the same selected model. 
We do not think that a general analytical description of this inference region is possible. 
We thus circumvent this problem by using a Monte Carlo approximation, adapting and extending the existing approaches summarized in Section~\ref{sec:infph}.

\subsection{Powerful Inference for $L_2$-Boosting with Linear Base-learners} \label{subsec:adaption2}

We base inference on the potentially multiply truncated Gaussian distribution of $\tilde{R} = \bm{v}^\top \bm{Y}$ conditional on 
$\bm{P}_v^\bot \bm{y}$ and the selection $\tilde{R} \in \mathcal{R}_y$. Then, the truncated normal density of $\tilde{R}$ is given by 
\begin{equation}
f(R) \propto \exp \left\{ -\frac{1}{2\sigma^2 \bm{v}^\top\bm{v}} (R - \bm{v}^\top \bm{\mu})^2 \right\} \cdot  \mathbbm{1} \{ R \in \mathcal{R}_y \}, \label{tnd}
\end{equation}
where $\mathcal{R}_y$ is a union of polyhedra. The proof of equation (\ref{tnd}) follows analogously to Lemma 1 of \citet{Yang.2016} for $\tilde{R} = \bm{v}^\top \bm{Y}$ using $w=1$ (cf. (\ref{eq:yangDens})). Note that $\langle \text{dir}_{\bm{W}}(\bm{y}), \bm{\mu} \rangle = \bm{v}^\top \bm{\mu} / ||\bm{v}||_{2}$ in this case; we rescaled $\tilde{R}$ compared to the definition before and kept the sign by using a normal instead of a $\chi$-distribution. Let $r_\text{obs} = \bm{v}^\top \bm{y}$. Then, analogous to \citet{Yang.2016} we can define a p-value by $$\varsigma(\beta_{j,0}) = \frac{\int_{\tilde{R} > r_\text{obs}, \tilde{R} \in \mathcal{R}_y} e^{-(2\sigma^2 \bm{v}^\top \bm{v})^{-1} (\tilde{R}^2 - 2\tilde{R}\beta_{j,0})} \, \mathrm d\tilde{R}}{\int_{\tilde{R} \in \mathcal{R}_y} e^{-(2\sigma^2 \bm{v}^\top \bm{v})^{-1} (\tilde{R}^2 - 2\tilde{R}\beta_{j,0})} \, \mathrm d\tilde{R}}$$ for $H_0: \bm{v}^\top\bm{\mu} = \beta_{j,0}$ and since the truncated Gaussian distribution with potentially multiple truncation limits increases monotonically in its mean $\rho$ \citep[see, e.g.,][]{Ruegamer.2018b}, we can find unique values $\rho_{\alpha/2}, \rho_{1-\alpha/2}$ for any $\alpha \in (0,1)$, such that $$\varsigma(\rho_a) = \frac{\int_{\tilde{R} > r_\text{obs}, \tilde{R} \in \mathcal{R}_y} e^{-(2\sigma^2 \bm{v}^\top \bm{v})^{-1} (\tilde{R}^2 - 2\tilde{R}\rho_a)} \, \mathrm d\tilde{R}}{\int_{\tilde{R} \in \mathcal{R}_y} e^{-(2\sigma^2 \bm{v}^\top \bm{v})^{-1} (\tilde{R}^2 - 2\tilde{R}\rho_a)} \, \mathrm d\tilde{R}} = a,$$ $a \in \{ \alpha/2, 1-\alpha/2 \}$, to construct a two-sided confidence interval $[\rho_{\alpha/2},\rho_{1-\alpha/2}]$. This is an extension of the one-sided confidence intervals of \citet{Yang.2016}. 

Note that $\varsigma(\rho_a)$ can then be rewritten as 
\begin{equation}
\frac{\mathbb{E}_{\tilde{R}\sim\mathcal{N}(0,\sigma^2 \bm{v}^\top\bm{v})} \left[  \mathbbm{1} \{ \tilde{R} \in \mathcal{R}_y, \tilde{R} > r_\text{obs}  \} \cdot e^{(\sigma^2 \bm{v}^\top \bm{v})^{-1} \tilde{R}\rho_a} \right]}{\mathbb{E}_{\tilde{R}\sim\mathcal{N}(0,\sigma^2 \bm{v}^\top\bm{v})} \left[  \mathbbm{1} \{ \tilde{R} \in \mathcal{R}_y \} \cdot e^{(\sigma^2 \bm{v}^\top \bm{v})^{-1} \tilde{R}\rho_a} \right]}, \label{eq:papprox}
\end{equation}
which allows for an empirical approximation as in (\ref{yangapprox}). Further note that this approach does not require to condition on the list of signs nor on the order of the selected variables. As this has been necessary to obtain selective inference statements for existing approaches such as the Lasso \citep{Lee.2016}, our method can also be used to condition on less for these approaches and thus potentially leads to more powerful inference.\\

\subsubsection*{Monte Carlo Approximation}

In practice, importance sampling from $\Pi = \mathcal{N}(r_\text{obs}, \allowbreak \sigma^2 \bm{v}^\top\bm{v})$ approximates (\ref{eq:papprox}) well if the given truncation limits around $r_\text{obs}$ are fairly symmetric, yielding the weights $w_b = \exp((2r^b r_\text{obs} - r_\text{obs}^2) \cdot (-2\sigma^2 \bm{v}^\top\bm{v})^{-1})$ for the importance sampler. A refinement of the sampling routine is necessary to also work well in more extreme cases. An example frequently encountered in practice is when $r_\text{obs}$ is rather large and at the same time lies very close to one truncation limit, yielding an insufficient number of samples $r^b \in \mathcal{R}_y$ to approximate the tail of the truncated distribution well. We therefore propose a more efficient sampling routine, motivated by and applicable to selection procedures, for which the support of the truncated distribution is known to be a single interval $[\mathcal{V}^{lo},\mathcal{V}^{up}]$. Our idea is that, in this case, we do not need to characterize the space empirically since the distribution of interest is known with the exception of the interval limits (the variance is assumed to be known and the null distribution determines the mean $\rho$). By employing a line search, we can find $\mathcal{V}^{lo},\mathcal{V}^{up}$ and conduct inference based on the truncated normal distribution function $F_{\rho,\sigma^2 \bm{v}^\top\bm{v}}^{[\mathcal{V}^{lo},\mathcal{V}^{up}]}(\cdot)$. We use such a corresponding line search here to refine the importance sampling. To find a super set of $\mathcal{R}_y$, we start with extremely small, or respectively, large quantiles $R$ of $\Pi = \mathcal{N}(\rho,\sigma^2 \bm{v}^\top \bm{v})$ and check for selection congruency, i.e., whether $R \in \mathcal{R}_y$. We successively increase, or respectively, decrease the quantiles for which we perform a congruency check if the corresponding values are not in $\mathcal{R}_y$ until they are, and choose $\widetilde{R}^{lo}, \widetilde{R}^{up}$ as the last values outside $\mathcal{R}_y$. This gives a superset of the support of $R$ up to numerical precision using the order of 50 refits of the model. We then draw from a uniform distribution with support $[\tilde{R}^{lo},\tilde{R}^{up}]$. In comparison to sampling from $\Pi$, finding preliminary truncation limits $[\tilde{R}^{lo},\tilde{R}^{up}]$ to refine the sampling space prior to sampling notably enhances accuracy and efficiency due the increased number of accepted samples. 


The number of samples required to sufficiently approximate the expectations in (\ref{eq:papprox}) depends on the approximation quality of the importance sampling. The crucial point here is the representative nature of samples that are required to draw from $\Pi$ in order to get the same efficiency as given by the estimator based on samples from $\mathcal{U}[\tilde{R}^{lo},\tilde{R}^{up}]$. This can be examined by estimating the effective sample size $n_e$, which represents the number of samples that we are required to draw from $\Pi$ in order to obtain the same efficiency as using the estimator based on the given number of samples from $\mathcal{U}[\tilde{R}^{lo},\tilde{R}^{up}]$. Practitioners can evaluate this by estimating $n_e$ using $\hat{n}_e = (\sum_{b=1}^B w_b)^2 / (\sum_{b=1}^B w_b^2)$ \citep[see, e.g.,][]{Martino.2017}. 
In order to set an appropriate number of samples, this information can be used to assess the Monte Carlo error and choose the number of samples based on the desired approximation quality. A more pragmatic solution is to increase the number of samples gradually until the resulting inference statements do not noteably change.

\subsection{Further extensions} \label{extensions}

The ideas in Section \ref{subsec:adaption1} and \ref{subsec:adaption2} can be extended to allow for computations in further relevant settings. 
We discuss four practically important extensions.\\

\noindent \emph{Inference for groups of variables}. In order to test groups of variables, the approach by \citet{Yang.2016} described in Subsection~\ref{sec:infph} can almost directly be applied. To this end, we define $\mathcal{S}$ based on the set of chosen variables and use the sampling approach proposed in Subsection~\ref{subsec:adaption2} for the $\chi$-distribution on $\mathbb{R}^{+}$, such that $\tilde{R}^{lo} \geq 0$.\\

\noindent \emph{Incorporating cross-validation and other sub-sampling techniques}. One of the most common ways to choose a final stopping iteration $m_{\text{stop}}$ for the boosting algorithm is by using a resampling technique such as $k$-fold cross-validation (CV) and estimating the prediction error of the model in each step. By choosing the model with the smallest estimated prediction error, we again exploit information from the data, which we have to discard in the following inference. $\mathcal{S}$ then corresponds to the selection obtained using $L_2$-Boosting with stopping iteration chosen by CV. We can extend the sampling approach described in Section~\ref{subsec:adaption2} by incorporating the CV conditions into the space definition of $\mathcal{R}_y$. Define a (multivariate) random variable $\bm{\Delta}$ describing these conditions, which is independent of $\bm{Y}$. For $k$-fold CV, for example, $\bm{\Delta}$ is a uniformly distributed random variable on all possible permutations of $(1,\ldots,1,2,\ldots,2,\ldots,k,\allowbreak\ldots,k)$, yielding the assignments $\bm{\delta} = (\delta_1,\ldots,\delta_n)$ for every entry in $\bm{y}$ to one of the $k$-folds with equal probability (if $n$ is a multiple of $k$). To conduct inference, we additionally condition on $\bm{\Delta} = \bm{\delta}$, i.e., we keep the folds fixed and identical to those of the original fit, when rerunning the algorithm with a new sample $\bm{y}^b$ to check for consistency with the observed selection event $\mathcal{R}_y$. In fact, this approach is not only restricted to resampling methods. Stability selection \citep{Shah.2013} or other possibilities to choose an ``optimal'' number of iterations, as for example, by selection criteria such as the Akaike Information Criterion \citep[AIC,][]{Akaike.1974} can be incorporated into the inference framework in the same manner. 
For a mathematical justification observe that conditional on the selection event $\mathcal{R}_y$ (including conditions on other random variables such as $\bm{\Delta}$), $\bm{P}_{\bm{W}}$ is fixed and Lemma 1 by \citet{Yang.2016} holds analogously.\\ 

\noindent \emph{Unknown error variance}. If the true error variance is unknown, we may use a consistent estimator instead. Judging by our simulation results, the effect of plugging in the empirical variance of the boosting model residuals is negligible in many cases and may also be a better (less anti-conservative) choice than the analogous estimator given by an ordinary least squares estimation in the selected model due to the shrinkage effect. In cases with smaller signal-to-noise ratio, however, the plug-in approach may also yield invalid p-values under the null as shown in our simulation section. \citet{Tibshirani.2015} present a plug-in as well as a bootstrap version of the test statistic, which yield asymptotically conservative p-values. The bootstrap approach, however, can only be conducted efficiently if truncation limits of the test statistic are known. In the simulation section, we investigate the first suggestion by \citet{Tibshirani.2015} -- using the empirical variance of $\bm{y}$ as a conservative estimate for $\sigma^2$ -- which better suits the presented framework.\\

\noindent \emph{Smooth effects}. The presented approach can also be used for additive models when the linear predictor $\eta_i = \bm{x}_i^\top \bm{\beta}$ in the working model $y_i = \eta_i + \varepsilon_i, i = 1,\ldots,n$ is extended by additive terms of the form $g(c_i)$ for some covariate $\bm{c} = (c_1,\ldots,c_n)^\top$. For ease of presentation, we assume that only one covariate $\bm{c}$ is incorporated with an additive term, but the general case is analogous. We use a basis representation $g(c_i) = \bm{B}(c_i)\bm{\gamma} = \sum_{\upsilon=1}^M \bm{B}_\upsilon(c_i) \gamma_\upsilon$ with $M$ basis function $B_\upsilon(\cdot)$ evaluated at the observed value $c_i$, basis coefficients $\gamma_\upsilon$, $\bm{B}(c_i) = (B_1(c_i), \ldots, B_M(c_i))$ and $\bm{\gamma} = (\gamma_1,\ldots,\gamma_M)^\top$. 
We are interested in testing the best linear approximation of $\bm{\mu}$ in the space spanned by a given design matrix $\bm{X}_{\mathcal{A}}$, where $\bm{X}_{\mathcal{A}}$ now, not only contains all selected variables with linear effect, but also the columns $\widetilde{\bm{B}} = ({\bm{B}}(c_1)^\top,\ldots,{\bm{B}}(c_n)^\top)^\top$ with the basis functions evaluated at $\bm{c}$. In particular, we may want to perform a point-wise test $H_0 : \mathfrak{g}(c) = 0$ for some $c$, where $\mathfrak{g}$ is the ``true'' function in the basis space resulting from the best linear approximation of $\bm{\mu}$ by the given model. $H_0$ can be tested using the proposed framework with test vector $\bm{v}^\top = {\bm{B}}^0(c) (\bm{X}_\mathcal{A}^\top\bm{X}_\mathcal{A})^{-1} \bm{X}_\mathcal{A}^\top$, as $\mathfrak{g}(c) = \bm{v}^\top \bm{\mu}$, where ${\bm{B}}^0(c)$ has the same structure as one row of $\bm{X}_{\mathcal{A}}$ but with all columns except those corresponding to $\bm{B}(c)$ set to zero. Instead of a point-wise test, the whole function can be tested
\begin{equation}
H_0: \mathfrak{g}(\cdot) \equiv \bm{0} \label{eq:fullfuntest}
\end{equation}
by 
regarding the columns in $\widetilde{\bm{B}}$ as groups of variables and setting $\bm{W}$ in (\ref{eq:yanghyp}) to $\bm{P}^\bot_{\bm{X}_{\mathcal{A}\backslash j}} \widetilde{\bm{B}}$, where $\bm{X}_{\mathcal{A}\backslash j}$ denotes $\bm{X}_\mathcal{A}$ without the $M$ columns of $\widetilde{\bm{B}}$.

The proposed tests and testvectors $\bm{v}$ or matrices $\bm{W}$ can also be used when smooth effects are estimated using a penalized base-learner with $\bm{D}_j \neq \bm{0}$. We note that this is one of the advantages of $L_2$-Boosting over the Lasso, as fitting smooth effects is not as straightforward for the Lasso.

\section{Simulations} \label{sec:simulation}

We now provide evidence for the validity of our method for linear and spline base-learners based on $B=1000$ samples per iteration and $\varrho = 1000$ simulation iterations. We also show the performance of the proposed method in comparison to the polyhedron approach in a relevant setting and investigate the effect of different variance values. For linear regression with linear base-learners the true underlying model is given by 
\begin{equation}
y_i = \eta_i + \varepsilon_i = \bm{X}_{[i,1:4]} \bm{\beta} + \varepsilon_i, \quad i=1,\ldots,n, \label{limoTrue1}
\end{equation}
where $\bm{\beta} = (4, -3, 2, - 1)^\top$, $\bm{\eta} = (\eta_1,\ldots, \eta_n)^\top$, $\varepsilon_i \overset{iid}{\sim} \mathcal{N}(0,\sigma^2)$ with $\sigma$ defined such that the signal-to-noise ratio $\text{SNR} := (\text{sd}(\bm{\eta})/\sigma) \in \{1,4\}$ and $[i,1:4]$ indicates row $i$ and columns 1 to 4 of $\bm{X}$, respectively. We construct four linear base-learners for the four covariates $\bm{x}_1,\ldots,\bm{x}_4$ in $\bm{X}_{[,1:4]}$ and additionally build $p_0 \in \{4,22\}$ base-learners based on noise variables for $n \in \{25, 100\}$ observations, where the columns in $\bm{X}$ are independently drawn from a standard normal distribution (empirical correlations range from $-0.53$ to $0.48$). Note that the case $p_o=22$ and $n=25$ constitutes a setting, in which $p > n$ holds. Figure \ref{fig:simResultsLinear} shows the observed p-values versus the expected quantiles of the standard uniform distribution for settings in which either the true model or a model larger than the true model with all four signal variables is selected. This corresponds to selection events, in which the null hypothesis (1) holds for $j > 4$ and thus p-values of inactive variables should exhibit uniformity given the selection event $\mathcal{A}$. The mixture of uniform $U[0,1]$ p-values when aggregating across selected models again results in $U[0,1]$ p-values. Results are given in Figure~\ref{fig:simResultsLinear} ($n=25$) and in Figure~\ref{fig:simResultsLinear2} in the Supplementary Material ($n=100$).

\begin{knitrout}
\definecolor{shadecolor}{rgb}{0.969, 0.969, 0.969}\color{fgcolor}\begin{figure*}[h!]
\includegraphics[width=\maxwidth]{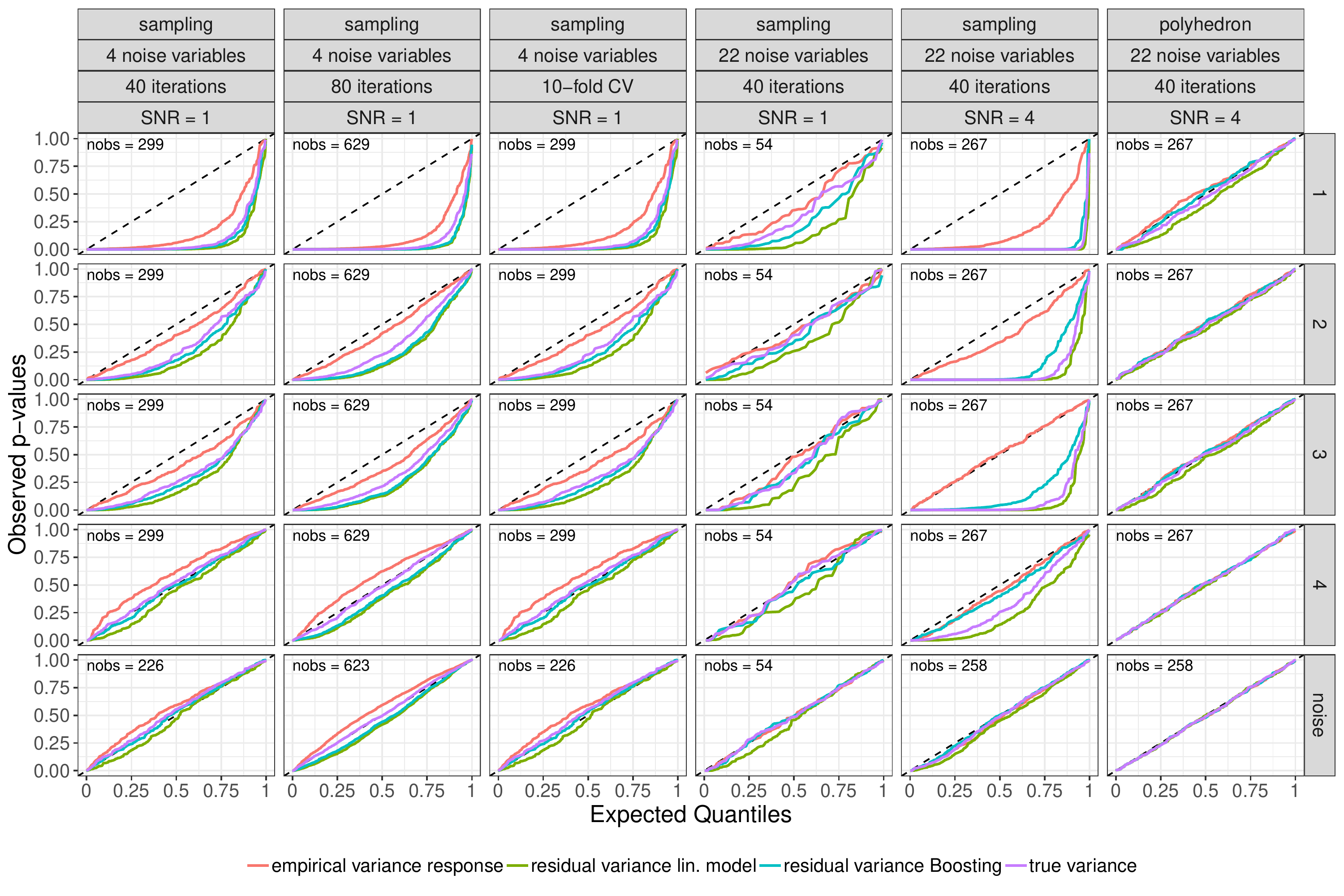} \caption[Observed p-values vs]{Observed p-values vs. expected quantiles across different covariates (rows) as well as different methods, number of noise variables, number of boosting iterations and SNR (columns) after boosting with a step-length of 0.1 using different variance types (colours), B = 1000, and a total of $\varrho = 1000$ simulation iterations in settings with n = 25. p-values are shown for simulation iterations, in which either the true model or a model larger than the true model is selected. For each setting, the number of those contributing iterations (nobs) out of $\varrho$ is noted in the left upper corner.}\label{fig:simResultsLinear}
\end{figure*}

\end{knitrout}

\noindent \emph{Results}: p-values for effects of ``true effect'' variables show deviations from the angle bisecting line, indicating the ability of the proposed procedure to correctly infer the significance of the effects. The power decreases for a smaller number of observations (cf. Figure~\ref{fig:simResultsLinear2}), a smaller SNR and a larger number of noise variables. Note that 22 noise variables here corresponds to a $p>n$-setting. The polyhedron approach yields correct p-values under the null, but shows no power for non-noise variables. p-values for the proposed approach (``sampling'') show much greater power. They are uniform under the null when using the true variance (even when selecting $m_\text{stop}$ using CV), with more conservative results when using the empirical variance of the response and slightly non-uniform p-values when using a plugin estimator. Differences are similar for larger $n$. In this respect, the empirical variance of boosting residuals is more favourable than that of an OLS refit, but can also lead to deviations. However, note that the empirical approximation of p-values is not very accurate in the settings where specific selection events are rather unlikely, as only a small number of samples $r^b \in \mathcal{R}_y$ can be used. These are typically the settings which also have small nobs. This could be improved by increasing the number of samples $B$. Our main findings can thus be summarized as follows: We conclude that our method produces valid inference, even without knowledge of the true variance by plugging in the empirical variance of the boosting residuals or a conservative estimate. Our approach is furthermore able to detect small effects in high-dimensional settings and / or settings with a larger signal-to-noise ratio and can successfully be extended to include sub-sampling schemes in selective inference statements.


Corresponding confidence intervals of the proposed test procedure reveal approximately $(1-\alpha)$\% coverage for the same simulation settings. Results for $\alpha = 0.05$ are given in Table~\ref{covTab}. Deviations from the ideal coverage of $95$\% are primarily due to numerical imprecision when inverting the hypothesis test and more accurate results can be obtained in applications when the number of non-rejected samples is too low by increasing the number of samples $B$.

\begin{table}[ht]
\centering
\begin{small}
\caption{Estimated coverage of selective confidence intervals obtained by the proposed sampling approach for $n=25$ observations when using the true variance in different settings (columns) in which either the true model or a model larger than the true model is selected.} 
\begin{tabular}{rccccc}
\cline{2-6}
 & \multicolumn{5}{c}{\underline{$p_0$, number of iterations, SNR}} \\ 
&  4,40,1 &  4,80,1 & 4,CV,1 & 22,40,1 & 22,40,4 \\ \hline
noise & 0.9566 & 0.9571 & 0.9618 & 0.9485 & 0.9211 \\ 
signal & 0.9699 & 0.9559 & 0.9326 & 0.9444 & 0.9429 \\ 
   \hline
\end{tabular}
\label{covTab}
\end{small}
\end{table}

In the Supplementary Material, we additionally provide results for other settings of this simulation study as well as results for additive models using spline base-learners. Here the true underlying function is given by $y_i = \text{sin}(2 X_{[i,1]}) + \frac{1}{2}X_{[i,2]}^2 + \varepsilon_i, i=1,\ldots,n=300$, $\varepsilon_i \overset{iid}{\sim} \mathcal{N}(0,\sigma^2)$ with $\sigma$ defined such that the signal-to-noise ratio $\text{SNR} = 0.5$, and 13 further covariates $\bm{X}_{[,3:15]}$. All covariate effects are represented using penalized B-splines \citep[P-splines;][]{Eilers.1996} with B-Spline basis of degree 3, 5 knots and second order differences penalty. Tests for the whole function are performed as proposed in (\ref{eq:fullfuntest}). Results suggest very high power but uniformity of p-values for noise variables, supporting the conclusion that the proposed test also works well for additive terms. 

We further compare the selective approach for linear base-learners with the naive approach, thereby illustrating the invalidity of classical unadjusted inference (see Figure~\ref{fig:simResultsLinear2}), compare the length of selective and naive intervals (Figure~\ref{fig:lengthComp}) and address the criticism of potentially infinite selective intervals. Investigating the frequency of an infinite interval for two simulation scenarios for $n=100$ and $p = 26$ (Figure~\ref{fig:infLength}) shows that inifinite length of corresponding intervals occurs only in around 5\% of all cases. 


\subsection{Computation time and further details}

As the proposed framework requires refitting the selection procedure $B$ times, the computation time might be the biggest concern for practitioners. When it is not possible to parallelize the model fits for the values $r^b$, increasing $B$ obviously results in a linear increase of computation time similar to conducting a boostrap. In comparison to the model refits, the preceding line search for the limits of $\mathcal{R}_y$ can be rather cheap, but may take a predominant amount of time if the selection event $\mathcal{S}(\bm{Y}) = \mathcal{A}$ has a very small probability for the given (latent) data generating process. This can, e.g., result in a highly fragmented support and / or very small selection regions, making a proper line search and approximation rather tedious. For these rare events, practitioners have the choice to either avoid extended run-times by using a sampling approach without a preceding search for the limits of $\mathcal{R}_y$ or to obtain more accurate inference results by using the line search approach with additional run-time. We note, however, that without a preceding line search, sampling may yield a very small number of un-rejected samples and low accuracy of inference statements in this case. In order to give a rough insight into run-times for our software, we provide computation times for the sampling itself using different settings for $n$ and $p$. These include realistic, high-dimensional setups after model selection with subsequent $5$-fold CV. Estimated run-times with parallelization of the $5$-fold CV but without parallelization of the refitting procedure itself are shown in Figure~\ref{fig:compTime} in the Supplementary Material D for inference statements on one hypothesis (one projection direction) based on $\varrho = 5$ replications per setting and $B=1000$. Results suggest that computation time is sublinear in $n$, which is due to the fact, that the hat matrix will only be computed once for all refits, but computing time for fixed $n$ seems to roughly increase as $\mathcal{O}(p^2 \log(p))$. 

Although the sampling approach has a larger than linear effort in $p$, we note that for our largest simulated setting, computation can be done in less than a day when parallelizing on 25 cores. By contrast the polyhedral approach suffers from a memory problem, as calculations involve the storage of and matrix operations on the $(2 \cdot (p - 1) \cdot m_\text{stop} ) \times n = 222000 \times 10000$ matrix $\bm{\Gamma}$, which when stored as a vector, exceeds the theoretical limit of elements in R. A possible solution to this bottleneck would need to distribute the matrix as well as computations on it across different cores.


\section{Application} \label{sec:application}

We now apply our framework to a data set for the prediction of sales prices of real estate single-family residential apartments in Tehran, Iran. The data set includes 372 observations and 105 continuous covariates including 19 economic variables, such as the amount of loans extended by banks in a quarter or the official exchange rate with respect to dollars (with 5 different lags for each variable) and 8 physical / financial variables, such as the duration of construction, the total floor area of the building or the preliminary estimated construction cost of the project. The data set has previously been analyzed by \citet{Rafiei.2015} and is freely available in the UCI Machine Learning data set repository (\url{https://archive.ics.uci.edu/ml/datasets/}). We use a flexible additive working model with 7 factor variables (piecewise constant interest rates and the location of the building) as well as 93 metric variables and check the linearity assumption of all covariates by additionally including 93 non-linear deviations from the linear effects. In order to estimate the smooth effects, we fit the model using cubic P-spline base-learners with second-order difference penalties and 7 knots per spline. Our full model thus corresponds to a $p>n$-setting. Splitting effects into a linear effect and a non-linear deviation from the corresponding linear effect also facilitates a fair base-learner selection in boosting \citep{Hofner.2011}. The optimal stopping iteration $m_\text{stop} = 149$ for the boosting algorithm with step-length $\nu = 0.1$ is found by using 5-fold cross-validation, which is incorporated into the selection mechanism $\mathcal{S}$. After $149$ iterations, five non-linear effects (three physical / financial and two economic variables) and 11 linear effects (3 physical / financial, 7 economic variables and the starting year of constructions) are selected by the boosting procedure. The non-linear deviations show a U- or inverse U-shape, which is shown in the Supplementary Material. We use the proposed sampling approach with $B=1000$ samples, separately testing linear effects using (11) and testing non-linear deviations as in (12). This yields a significant linear as well as non-linear effect of the square meter price at the beginning of the project, a significant non-linear effect for the population size of the city, and significant linear effects of the duration of construction, the number of loans extended by banks and the unofficial exchange rate with respect to dollars. All other effects are found not to be significant at a 5\%-level. In comparison, a standard linear model including all covariates, yields three further significant physical variables (project locality, lot area and a preliminary estimate of construction costs) and six additional significant economic variables with different lags. In contrast to the boosting approach with subsequent inference, more significant variables are found by the standard inference procedure as no information in the data is used for model selection. This, however, restricts the additive model to linear effects only. In addition, standard software automatically excludes 29 of the economic variables due to collinearity of the predictors.

\section{Discussion} \label{sec:extensions}

In this paper we propose an inference framework for $L_2$-Boosting by transfering and adapting several recently proposed selective inference frameworks. As far as we know, there are no previous general methods available to quantify uncertainty of boosting estimates (or more generally for slow learners) in a classical statistical manner when variable selection is performed. 
Available permutation tests \citep{Mayr.2017b} are restricted to certain special cases and the conventional bootstrap cannot yield confidence intervals with proper coverage due to the bias induced by the shrinkage effect. 
We propose tests and confidence intervals for linear base-learners as well as for group variable and penalized base-learners. Using Monte Carlo approximation for the calculation of p-values and confidence intervals, we avoid the necessity for an explicit mathematical description of the inference space. This allows us to condition on less, which in turn increases power notably in comparison to polyhedron approaches. 

Selective inference can yield unstable and potentially infinite confidence intervals in certain situations. This was recently shown by \citet{Kivaranovic.2018} for selective inference concepts based on polyhedral constraints. However, for our method exploiting the fact that the selective space is a union of polyhedra, this seems to be rarely the case. Our simulation studies show powerful inference despite settings with a low signal-to-noise ratio and/or with the number of predictors exceeding the number of observations prior to model selection. This suggests that using the same approach for the Lasso selection when not conditioning on a list of signs or the variable order, which also results in a union of polyhedra, might help in obtaining more powerful inference.

We apply our framework to sales prices of real estates and, in contrast to existing approaches that combine model selection and subsequent inference, allow for non-linear partial effects as well as the selection of the stopping iteration using CV. Using simulation studies with a range of settings, we verify the properties of our approach.


This work opens up a variety of future research topics. In order to leave more information for inference and further reduce the occurence of infinite confidence intervals, the framework could be extended by incorporating randomization in the model selection and inference step \citep[see, e.g.][]{TianHarris.2016}. Adapting this concept for the given framework is, however, not straightforward as it is not clear whether estimators obtained by the boosting procedure are the solution to a closed-form optimization problem.


An extension to generalized linear models (GLMs) would be relevant but challenging since conditions involving $\bm{y}$ might imply conditioning on $\bm{y}$ itself if the response is discrete 
\citep[see][for more details on selective inference for GLMs]{Fithian.2014}. It would also be interesting to investigate whether the asymptotic results of \citet{Tian.2017} can be used to construct inference for CFGD algorithms other than $L_2$-Boosting.


\vskip 0.2in
\bibliography{mybibliography}

    
\clearpage    
\onecolumn

\bigskip
\begin{center}
{\large\bf Supplementary Material}
\end{center}
\section*{Supplementary Material A: Further Simulation Results}\label{appA}

\subsection*{A.1 Further Simulation Results for Linear Base-learners}

We first investigate the validity of our inference approach in two additional settings for $n=100$ observations. The results are visualized in Figure~\ref{fig:simResultsLinear2}, suggesting powerful and valid inference if the selective approach is used and proving the invalidity of classical inference (naive) when not adjusted for model selection. 

\begin{knitrout}
\definecolor{shadecolor}{rgb}{0.969, 0.969, 0.969}\color{fgcolor}\begin{figure}[h!]
\includegraphics[width=\maxwidth]{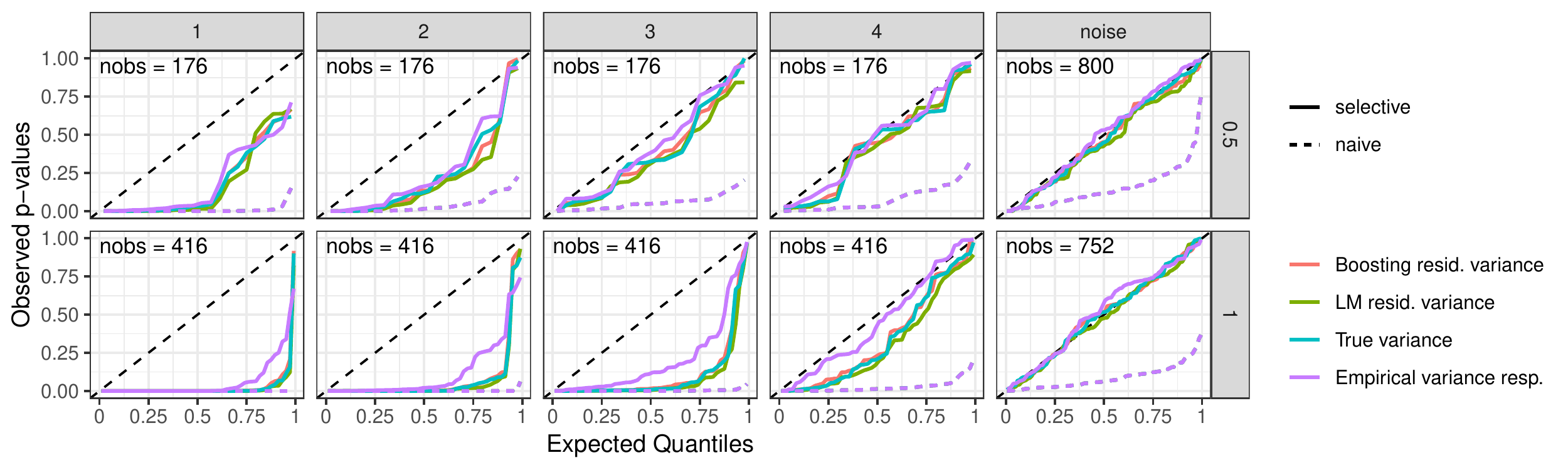} \caption[Observed p-values vs]{Observed p-values vs. expected quantiles across different covariates (columns) as well as different SNR (rows) for boosting with different variance values / estimates (colours), 26 variables including 22 noise variables, $B=1000$, a total of $\varrho = 1000$ simulation iterations and $n=100$ (in contrast to $n=25$ in the main article). p-values are shown for simulation iterations, in which either the true model or a model larger than the true model is selected. For each setting, the number of iterations (nobs) is noted in the left upper corner.}\label{fig:simResultsLinear2}
\end{figure}

\end{knitrout}

We further use the simulation scenario used for Figure~\ref{fig:simResultsLinear2} to examine the length of selective confidence intervals in comparison to naive confidence intervals (Figure~\ref{fig:lengthComp}) and investigate the frequency of observing an infinite length due to one or two infinite interval limits (Figure~\ref{fig:infLength}). Note that the given frequencies in Figure~\ref{fig:infLength} are an upper bound approximation since infinite interval limits can also occur due to the Monte Carlo approach with insufficient $B$ if not enough samples are congruent with the initial selection.

\begin{knitrout}
\definecolor{shadecolor}{rgb}{0.969, 0.969, 0.969}\color{fgcolor}\begin{figure}[ht]
\includegraphics[width=\maxwidth]{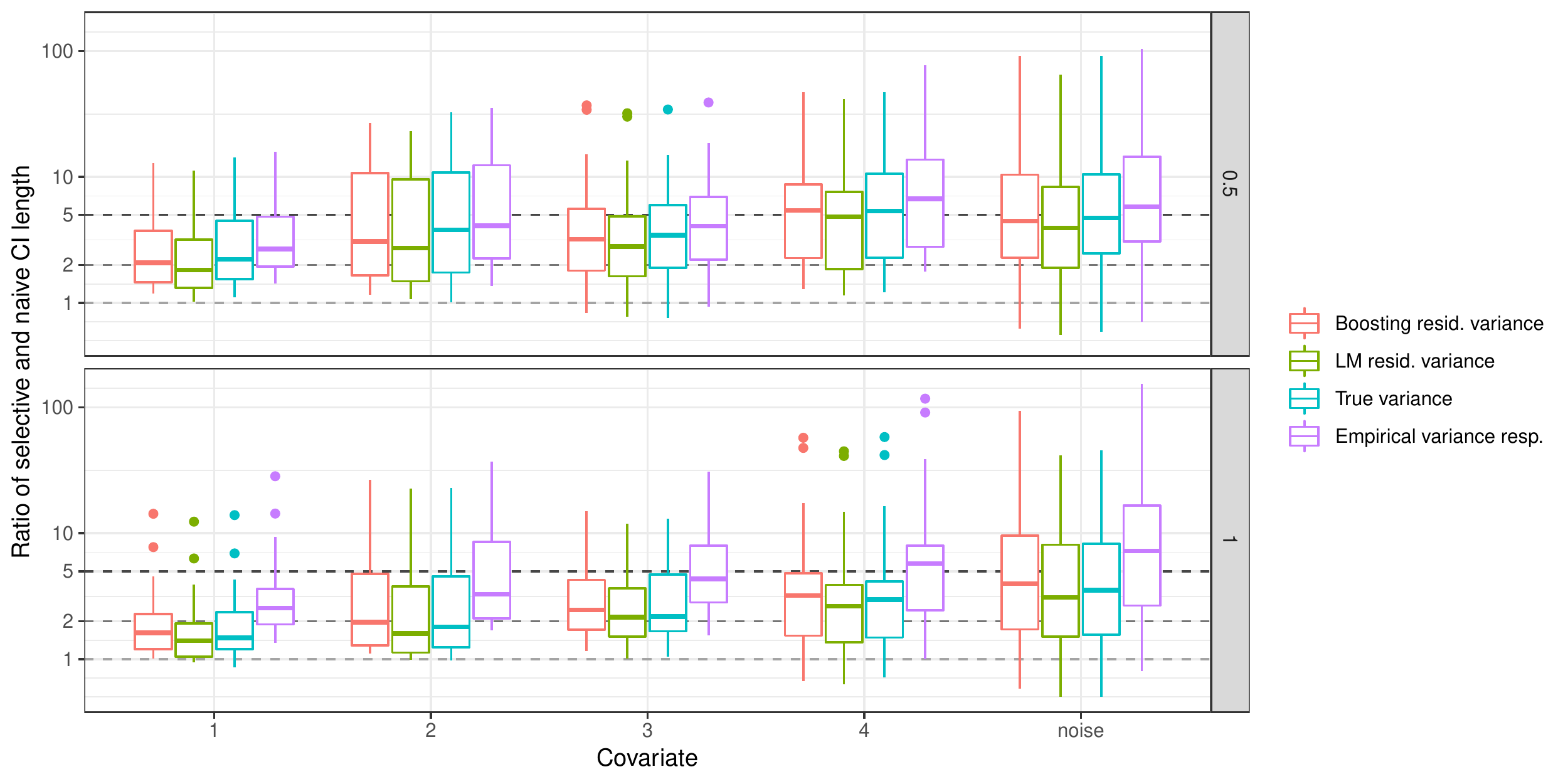} \caption[Ratio of selective confidence interval length divided by the classical confidence interval length for different SNR (rows) and variances (colours) used for the computation of the distribution of the test statistic]{Ratio of selective confidence interval length divided by the classical confidence interval length for different SNR (rows) and variances (colours) used for the computation of the distribution of the test statistic. Note that the y-axis is on a logarithmic scale.}\label{fig:lengthComp}
\end{figure}

\end{knitrout}

\begin{knitrout}
\definecolor{shadecolor}{rgb}{0.969, 0.969, 0.969}\color{fgcolor}\begin{figure}[ht]
\includegraphics[width=\maxwidth]{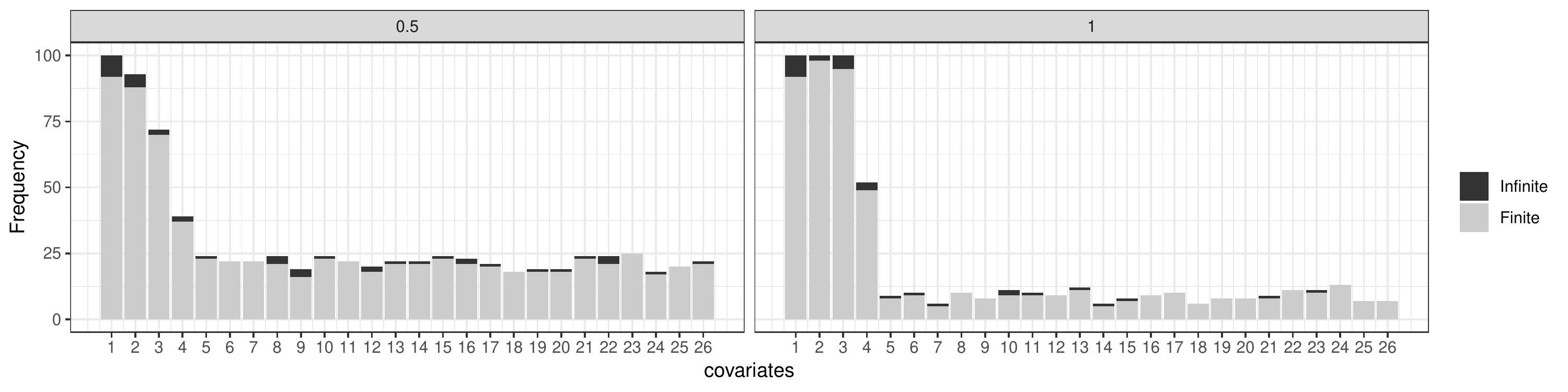} \caption[Frequency of finite / infinite interval lengths in two SNR settings (columns) for 100 simulation iterations]{Frequency of finite / infinite interval lengths in two SNR settings (columns) for 100 simulation iterations. Iterations, for which the corresponding variable was not selected, do not contribute to the bars. Variables 5 - 26 correspond to noise variables.}\label{fig:infLength}
\end{figure}

\end{knitrout}

\subsection*{A.2 Further Simulation Results for P-spline Base-learners}

Figure~\ref{fig:simResultsNonLinear} shows further simulation results for additive models as discussed in Section 5.

\begin{knitrout}
\definecolor{shadecolor}{rgb}{0.969, 0.969, 0.969}\color{fgcolor}\begin{figure}[ht]
\includegraphics[width=\maxwidth]{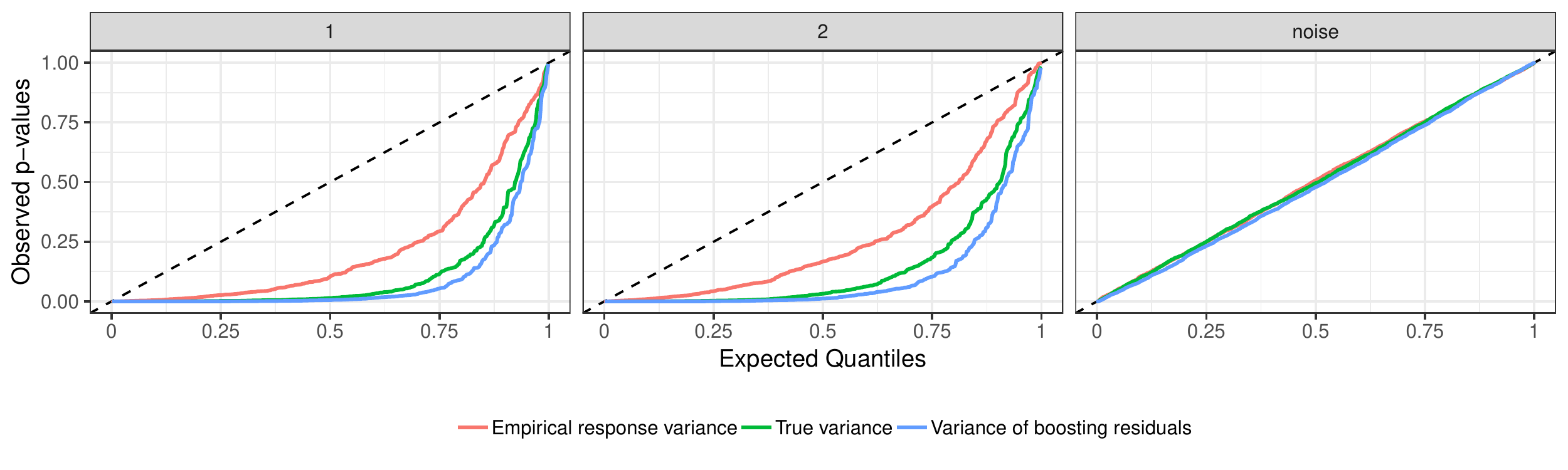} \caption[Observed p-values vs]{Observed p-values vs. expected quantiles across different covariates (columns) as well as different variance values / estimates (colours) for SNR = 1 for testing a function using boosted P-spline baselearners after 50 iterations and a step-length of 0.1, using a total of 500 simulation iterations. p-values are shown for simulation iterations, in which either the true model or a model larger than the true model is selected. All plots are based on 500 simulation iterations as the selection procedure always selected a model with both truly non-linear effects and (potentially) further noise variables.}\label{fig:simResultsNonLinear}
\end{figure}

\end{knitrout}


\section*{Supplementary Material B: Further Application Results}\label{appB}

The following plots visualize the estimated effects of the selected variables (after centering the variables) in the boosted additive model. The selected non-linear deviations are the total area of the building (physical variable 2), the lot area size (physical variable 3), the square-meter price of the unit at the beginning of the project (physical variable 8), the unofficial exchange rate with respect to dollars (economic variable 14) and the population of the city (ecnomic variable 18). 
Further selected variables (with linear effects) are the starting year of the project (START.YEAR), preliminary estimated construction cost based on the prices at the beginning of the project in a selected base year (physical variable 6), the duration of construction (physical variable 7), the number of building permits (economic variable 1), the number of loans extended by banks (economic variable 8) and the interest rate for loan (economic variable 10).

\begin{knitrout}
\definecolor{shadecolor}{rgb}{0.969, 0.969, 0.969}\color{fgcolor}\begin{figure}[h]
\includegraphics[width=\maxwidth]{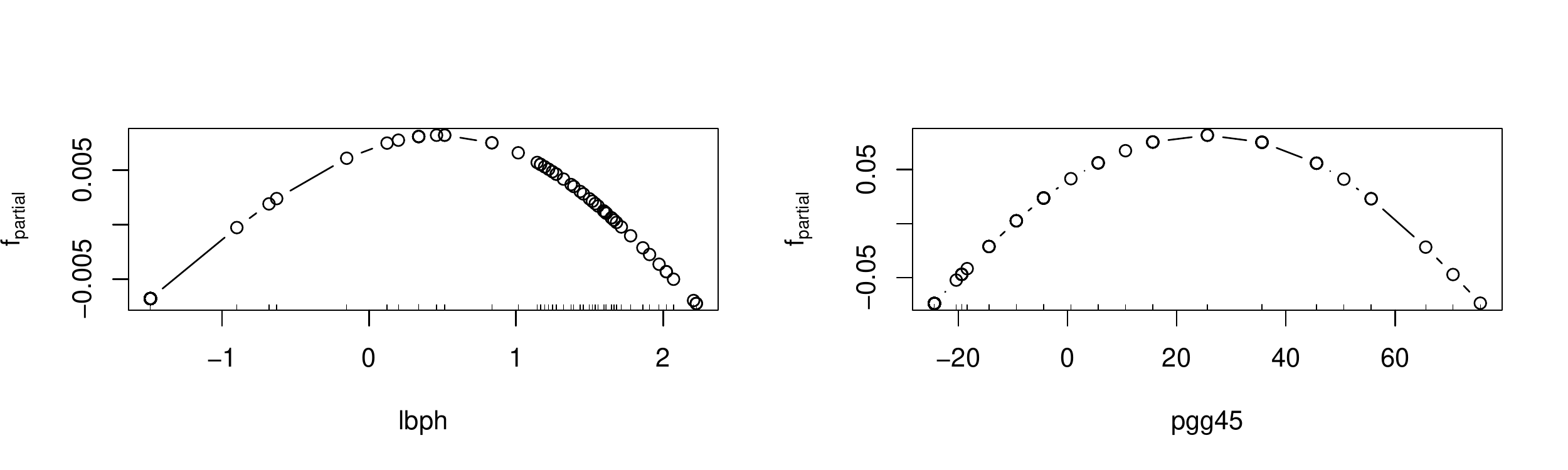} \caption[Partial effects of estimated linear and non-linear deviations for the selected covariates]{Partial effects of estimated linear and non-linear deviations for the selected covariates.}\label{fig:appl}
\end{figure}

\end{knitrout}

\section*{Supplementary Material C: Simulation Code}

The R-code and link to the software to reproduce simulation and application results can be found at \url{https://github.com/davidruegamer/inference_boosting}.

\section*{Supplementary Material D: Computation Time}

In the following an estimate of computation time of our software for different model setups is given. We use the same data generating process as in Section~\ref{sec:simulation}, assuming 4 signal variables and an SNR of $1$. Note that we did not use parallelization when sampling from the space $\mathcal{R}_y$ and run-times can be roughly divided by the number of cores, $\varphi$ when using parallelization over $\varphi$ cores. We use $B=1000$, $p_0 \in \{5, 50, 108\}$ noise variables and a grid from $1$ to $\min(p_0 \cdot 10^2, 10^4)$ iterations, in which the optimal stopping iteration $m_{\text{stop}}$ is searched for via CV.

\begin{knitrout}
\definecolor{shadecolor}{rgb}{0.969, 0.969, 0.969}\color{fgcolor}\begin{figure}[h]
\includegraphics[width=\maxwidth]{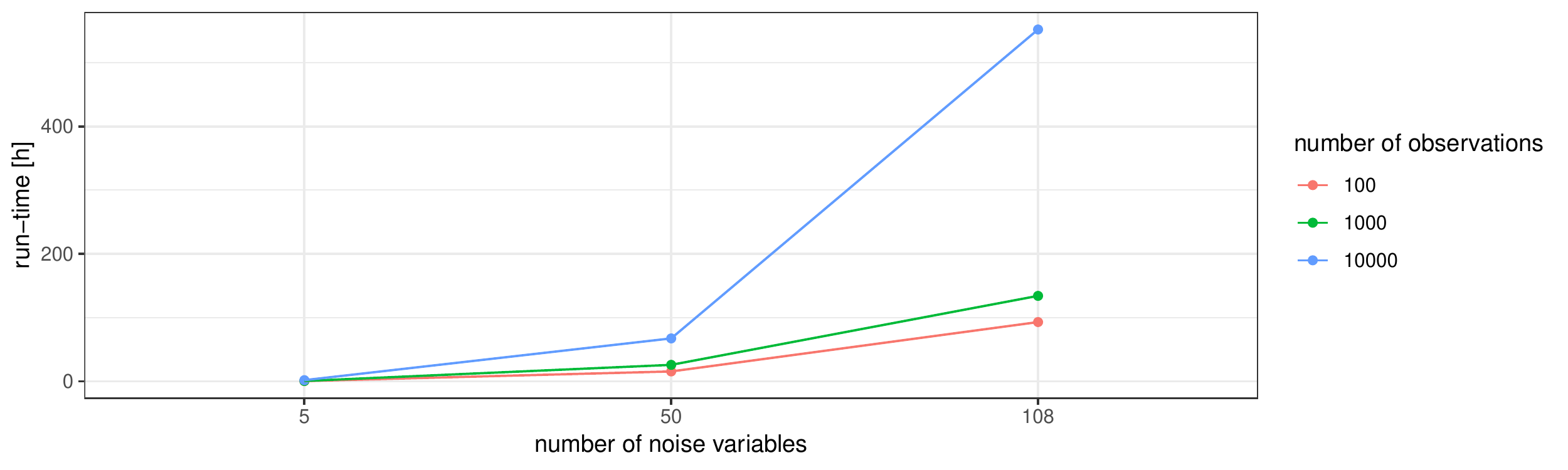} \caption[Average computation time in hours over 5 simulation iterations of our selective inference approach for one test vector and different numbers of noise variables (x-axis) as well as numbers of observations (colour)]{Average computation time in hours over 5 simulation iterations of our selective inference approach for one test vector and different numbers of noise variables (x-axis) as well as numbers of observations (colour).}\label{fig:compTime}
\end{figure}

\end{knitrout}

%


\bibliographystyle{spbasic}      


\end{document}